\documentclass[letterpaper, 10 pt, conference]{ieeeconf}  

\IEEEoverridecommandlockouts   

\usepackage{graphics} 
\usepackage{epsfig} 
\usepackage{mathptmx} 
\usepackage{times} 
\usepackage{amsmath} 
\usepackage{amssymb}  
\usepackage{color}
\usepackage{float} 
\usepackage{multirow}
\usepackage{subcaption}
\usepackage{makecell}

\usepackage{array}
\usepackage{makecell}

\title{\LARGE \bf Transformer-Based Warm-Starting for  Feasible and Optimal Terminal Approach to Tumbling Objects with Space Manipulators}

\author{Yuji Takubo$^{1*}$, Maximilian Adang$^{1*}$ 
, Mac Schwager$^{1}$, and  Simone D'Amico$^{1}$
\thanks{$^{1}$ Department of Aeronautics and Astronautics, Stanford University, Stanford, CA 94305. {\tt\small 
\{ytakubo, madang, schwager, damicos\}@stanford.edu}}
\thanks{*Equal contribution}
}

\begin{document}

\maketitle
\thispagestyle{empty}
\pagestyle{empty}

\begin{abstract}
Autonomous capture of a tumbling object with a space manipulator requires trajectory generation under real-time onboard constraints. 
However, coupled bus–arm dynamics, rotating safety geometry, and visibility requirements make the underlying optimization severely nonconvex, hence expensive to solve and sensitive to initialization.
This paper develops a causal-transformer warm-start for the computationally dominant attitude–manipulator stage of a two-stage sequential convex programming (SCP) framework. 
Across 300 held-out scenarios evaluated on embedded hardware, the learned warm-start reduces the iterations of the second-stage SCP by up to \(32\%\) and runtime by \(16\%\) while maintaining numerical feasibility and matching the final control cost. 
Furthermore, linear and flow matching action decoders are compared across action-chunk lengths and training dataset sizes, where the flow matching decoder exhibits stronger training-data efficiency than the linear decoder.
When SCP is terminated based on feasibility, the runtime is further reduced while the learned prior suppresses the severe high-cost tail observed with heuristic initialization. 
These results demonstrate that learned warm-starts can reduce onboard optimization cost while steering SCP toward basins containing feasible, low-cost solutions. 
\end{abstract}

\section{Introduction}

Space manipulators, consisting of a spacecraft equipped with a robotic arm, are regarded as key enablers for future space logistics operations, including on-orbit servicing of non-cooperative objects and in-orbit assembly and manufacturing.
However, these missions pose substantially greater challenges than conventional rendezvous missions to cooperative targets, primarily due to bus--arm dynamic coupling and the safety-critical nature of contact operations.

This work focuses on the terminal approach from a safe waypoint to the near-capture state, immediately before precise manipulator tracking of the capture point, under a \textit{free-flying} setting \cite{wilde2018equations} in which both the bus and manipulator are actuated.
In this regime, guidance is naturally formulated as a nonlinear programming (NLP)-based optimal control problem over the coupled bus--arm dynamics and safety constraints \cite{lampariello2010motion, virgili2019convex}.
However, these problems are highly sensitive to the initial guess, with both the optimality of the converged solution and the solver runtime strongly depending on warm-start quality.
Therefore, a high-quality warm-start is essential for onboard NLP-based motion planning.

Among various approaches for warm-start trajectory generation, learning-based sequence modeling has recently emerged as a promising direction \cite{art_ieeeaero24, celestini2024transformer, takubo2025towards, takubo2026agile}. 
These models autoregressively generate state--action sequences conditioned on task context, trajectory history, and desired outcomes \cite{chen2021decision}, and recent extensions further incorporate multimodal inputs \cite{black2024pi_0,takubo2026language}. 
Despite these advances, existing learning-based warm-starting methods have not been demonstrated for a realistic 3D terminal-approach scenario involving a space manipulator subject to highly nonlinear bus–arm dynamic coupling, tumbling-target attitude synchronization, rotating safety geometry, and field-of-view (FOV) cone constraints.
This combination makes full-state trajectory optimization expensive and makes heuristic initialization unreliable.

To this end, this paper presents a transformer-based rapid trajectory generation framework for a 3D space manipulator.
The contributions are summarized as follows:
\begin{itemize}
 \item A causal transformer (CT) warm-starting framework is developed for a 3D space manipulator terminal approach involving a tumbling target, coupled bus--arm dynamics, safety, and FOV cone constraints, solved via a two-stage sequential convex programming (SCP) pipeline.
 \item Linear and generative flow matching (FM) action decoders are systematically compared across action-chunk lengths and training dataset sizes to characterize their suitability for warm-starting this problem.
 \item Across 300 held-out scenarios, the learned warm-start reduces second-stage SCP iterations by up to \(32\%\) and runtime by \(16\%\) without degrading final cost, while suppressing the high-cost tail observed with heuristic warm-starting under feasibility-oriented SCP.
\end{itemize}

\section{Related Work} \label{sec:literature}

\subsection{Spacecraft Manipulator Motion Planning}

The guidance and control of space manipulators encompasses a broad range of problems, commonly organized by operational phase and by assumptions on spacecraft bus actuation \cite{flores2014review, papadopoulos2021robotic, hirano2018online}.
In particular, the terminal approach phase is often modeled as an NLP-based trajectory optimization problem \cite{virgili2019convex, lampariello2013generating} to ensure trajectory-wide feasibility, which cannot be guaranteed by a locally feasible safety filter \cite{loettgen2025control}. 
While a coupled bus--arm dynamics model is often adopted as the baseline \cite{wilde2018equations}, simultaneous optimization of full roto-translational bus actuation and joint motion is generally computationally prohibitive.
One way to reduce the problem size is to focus only on the manipulator torque and assume that the bus is unactuated \cite{lampariello2013generating}. 
Similarly, the terminal guidance of the e.Deorbit mission assumes perfect station-keeping of the spacecraft bus relative to the tumbling object's body frame \cite{telaar2017coupled}. 
An alternative approach is to plan the translational and rotational motion sequentially.
For example, \cite{virgili2019convex} adopts a two-stage motion-planning scheme that sequentially solves the translational trajectory planning of the center-of-mass (COM) and later solves the internal torque allocation between the bus and arm.

\subsection{Learned Warm-start for Trajectory Optimization}

Reliably generating a high-quality initial guess in real time for an NLP-based trajectory optimization is a recurring problem across aerospace and robotics. 
Existing approaches span a range of strategies, including look-up tables \cite{lampariello2013generating}, filter-based initializations \cite{yuan2025filtering}, direct prediction of complete trajectories \cite{banerjee2025deep}, and prediction of active or tight constraints \cite{briden2025transformer}. 
An alternative direction is to learn a policy that generates control actions from the problem context \cite{kim2022guided}. 
Within this class, sequence models that autoregressively predict actions or action chunks conditioned on tokenized trajectory context demonstrate promising performance for trajectory generation and optimization \cite{art_ieeeaero24, celestini2024transformer}. 
More recently, sequence model-based warm-starting has been applied to SCP for space manipulator trajectory generation \cite{d2026conditioned}, albeit only in a planar, low-dimensional setting with static obstacle avoidance. 
The present work extends this line of research to a 3D terminal-approach problem, in which the learned trajectory prior must initialize SCP under safety and FOV cone constraints attached to a tumbling target with coupled bus--arm dynamics.

\section{Preliminaries } \label{sec:prelim}

\begin{figure}[h!]
    \centering
    \includegraphics[width=0.99\linewidth]{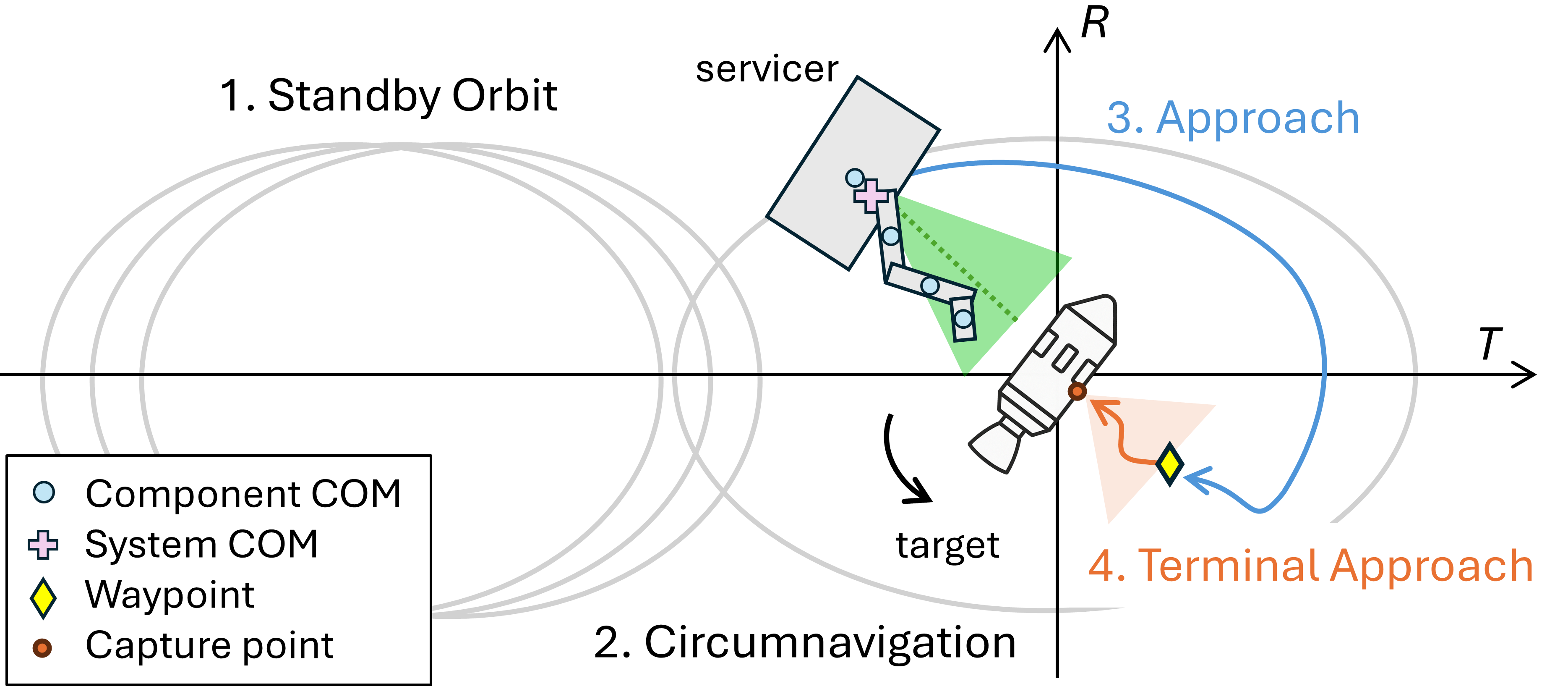}
    \caption{Concept of operations for the on-orbit servicing and capture of a tumbling object in the target-centered Radial--Tangential (RT) plane. }
    \label{fig:conops}
\end{figure}

This paper considers guidance and control for the terminal approach to an uncontrolled tumbling object using a space manipulator. 
In this work, the inertia properties, attitude, and tumbling rate of the uncontrolled object are assumed to be completely characterized; uncertainty in these quantities \cite{albee2021robust} is outside the scope of this paper. 
The concept of on-orbit servicing operations is summarized in Fig.~\ref{fig:conops}. 
First, the servicer approaches the tumbling target from a standby orbit (1) and enters a circumnavigation phase while maintaining safe separation (2). 
Upon completion of safe inspection, the servicer performs an approach toward the waypoint and synchronizes its attitude with the target (3). 
At this stage, passive safety may be compromised, and therefore an alternative collision avoidance strategy must be implemented. 
Finally, the terminal approach (4) is conducted from the waypoint a few meters away to the capture point. 
Until terminal approach, the servicer is usually modeled as a rigid body with the arm fixed; during terminal capture, the bus and arm are jointly actuated as a coupled multibody system.

\subsection{State Representation and Dynamics}

This work considers a spacecraft equipped with a serial manipulator composed of $N_l$ rigid links connected by 1-DOF revolute joints
\cite{wilde2018equations}. 
The generalized coordinates and generalized velocity are defined as
$ \boldsymbol{\eta}
    =
    [
        \mathbf{r}^\top, \mathbf{q}^\top, \boldsymbol{\alpha}^\top
    ]^\top
    \in \mathbb{R}^{7+N_l}, \
    \boldsymbol{\nu}
    =
[  \mathbf{v}^\top,  \boldsymbol{\omega}^\top , \dot{\boldsymbol{\alpha}}^\top
    ]^\top
    \in \mathbb{R}^{6+N_l},$
where $\mathbf{r}\in\mathbb{R}^3$ and $\mathbf{v}\in\mathbb{R}^3$ denote
the bus position and velocity in the target-centered Radial--Tangential--Normal (RTN) frame, $\mathbf{q}\in\mathbb{S}^3$ and $\boldsymbol{\omega}\in\mathbb{R}^3$ denote the bus attitude quaternion and angular velocity resolved in the target-centered RTN frame, and $\boldsymbol{\alpha}\in\mathbb{R}^{N_l}$ and $\dot{\boldsymbol{\alpha}}\in\mathbb{R}^{N_l}$ denote the joint angles and joint rates, respectively.

The coupled bus--arm dynamics are written as \cite{wilde2018equations}
{
\begin{subequations} \label{eq:eom_full_nonlin} 
\small 
\begin{align}
    \dot{\boldsymbol{\eta}} &= \mathbf{G}(\boldsymbol{\eta})\boldsymbol{\nu},\\
    \mathbf{H}(\boldsymbol{\eta})\dot{\boldsymbol{\nu}}
    + \mathbf{C}(\boldsymbol{\eta},\boldsymbol{\nu})\boldsymbol{\nu}
    &= \mathbf{u} + \mathbf{Q}_{\mathrm{orb}},
\end{align}
\end{subequations}
}
where $\mathbf{G}(\boldsymbol{\eta})$ is the kinematic mapping, $\mathbf{H}(\boldsymbol{\eta})$ is the generalized inertia matrix,
$\mathbf{C}(\boldsymbol{\eta},\boldsymbol{\nu})$ is the Coriolis/convective matrix, $\mathbf{u}= [\mathbf{f}^\top, \boldsymbol{\tau}_{0}^{\top}, \boldsymbol{\tau}_{m}^{\top}]^\top$ with $\mathbf{f}$, $\boldsymbol{\tau}_{0}$, and $\boldsymbol{\tau}_{m}$ denoting the commanded bus translational force, bus torque, and joint torque, respectively, and $\mathbf{Q}_{\mathrm{orb}}$ denotes the generalized forcing induced by the relative orbital dynamics. 

In implementation, $\dot{\boldsymbol{\nu}}$ is computed using a numerical forward-dynamics routine, without explicitly forming the closed-form equations of motion \cite{virgili2016spart}.
At each integration step, it computes the manipulator kinematics, maps external and control wrenches to the entire system, propagates articulated inertias and bias loads from the end-effector (EE) side to the bus, solves for the bus acceleration, and then recovers the joint accelerations through a forward recursion.
The Hill--Clohessy--Wiltshire (HCW) equation \cite{clohessy1960terminal} is used to compute the relative orbital acceleration, evaluated at each component COM and at the system COM (cf. Fig.~\ref{fig:conops}).
The differential accelerations are then converted into body wrenches and injected into the recursive dynamics, while the system COM acceleration is added directly to the bus translational acceleration.
The resulting continuous-time dynamics are discretized by forward Euler integration, yielding
{
\small
\begin{subequations}
\begin{align}
& \mathbf{x}_{k+1} = \mathbf{F}(\mathbf{x}_k, \mathbf{u}_k),  \label{eq:eom_discrete_nonlin} \\
& \mathbf{x}_k = [\boldsymbol{\eta}_k^\top, \boldsymbol{\nu}_k^\top]^\top \in \mathbb{R}^{13 + 2N_l}, \ \mathbf{u}_k = [\mathbf{f}_k^\top, \boldsymbol{\tau}_{0,k}^{\top}, \boldsymbol{\tau}_{m,k}^{\top}]^\top \in \mathbb{R}^{6 + N_l}. 
\end{align}
\end{subequations}
}

\section{Methodology}\label{sec:methodology}

This section presents the baseline two-stage SCP formulation for space manipulator trajectory optimization and the CT-based warm-starting strategy. 

\subsection{Nonconvex Trajectory Optimization}

The terminal approach trajectory generation can be cast as the following nonconvex trajectory optimization problem:
\begin{subequations} \label{eq:ocp}
\small
\begin{alignat}{2} 
    \min \quad &  {J}  \label{eq:ocp_obj}\\
    \text{s.t.} \quad 
    & \mathbf{x}_{k+1} = \mathbf{F}(\mathbf{x}_k, \mathbf{u}_k), \  
    && \forall k \in \mathbb{Z}_{1:N-1}, \ \label{eq:ocp_con_dyn} \\
    & g(\mathbf{x}_k, \mathbf{u}_k) = 0, \quad h(\mathbf{x}_k, \mathbf{u}_k) \leq 0, \quad   && \forall k \in \mathbb{Z}_{1:N},
\end{alignat}
\end{subequations}
where $J$ is the objective function, and $g(\mathbf{x}_k, \mathbf{u}_k)$ and $h(\mathbf{x}_k, \mathbf{u}_k)$ represent the possibly nonconvex equality and inequality constraints, respectively.
While one could solve a full-state coupled trajectory optimization over $ \{\mathbf{x}_k, \mathbf{u}_k \} \in \mathbb{R}^{19 + 3N_l}$, the resulting problem size is prohibitive for onboard use. 
Exploiting the dynamic decoupling between the system COM translation, driven solely by the external forces $\mathbf{f}_k$, and the internal motion about the system COM, this paper instead adopts the sequential two-stage scheme \cite{virgili2019convex}. 
Namely, Stage~1 plans the system COM translational trajectory, and Stage~2 resolves the internal bus and manipulator motion that realizes this trajectory while steering the EE to the target.
Although uncertainties in the target tumbling state and model parameters, as well as contingency requirements such as passive safety, can in principle be incorporated into Eq.~\eqref{eq:ocp} via formulations such as chance constraints \cite{takubo2025towards}, the present work assumes perfect state and model knowledge.

\vspace{0.25em}
\subsubsection{System COM translational trajectory optimization}

\normalsize

The decision variables for the system COM motion planning are defined as 
$\mathbf{x}_k^{(1)}
=
\begin{bmatrix}
\mathbf{r}_{c,k}^{\top} &
\mathbf{v}_{c,k}^{\top}
\end{bmatrix}^{\top}
\in\mathbb{R}^6,$
and
$
\mathbf{u}_k^{(1)}
=
\mathbf{f}_k
\in\mathbb{R}^3,
$
where $\mathbf{r}_{c,k},\mathbf{v}_{c,k}\in\mathbb{R}^3$ denote the system COM position and velocity in the target-centered RTN frame, and $\mathbf{f}_k$ is the net translational force. Stage 1 therefore solves
{
\footnotesize
\begin{subequations}
\begin{align}
\min \quad
& J_1 + J_2 \\
\mathrm{s.t.}\quad
& J_1
=
\sum_{k=1}^{N-1}
{\mathbf{u}_k^{(1)}}^{\!\top}
\mathbf{W}_1
\mathbf{u}_k^{(1)}, \\
& J_2
=
\sum_{k=1}^{N-2}
\left(
\mathbf{u}_{k+1}^{(1)}-\mathbf{u}_k^{(1)}
\right)^{\!\top}
\mathbf{W}_2
\left(
\mathbf{u}_{k+1}^{(1)}-\mathbf{u}_k^{(1)}
\right), \\
& \mathbf{x}_1^{(1)}
=
\mathbf{x}_{\mathrm{init}}^{(1)},
\qquad
\mathbf{x}_N^{(1)}
=
\mathbf{x}_{\mathrm{term}}^{(1)},
\label{eq:ocp_stg1_bc} \\
& \mathbf{x}_{k+1}^{(1)}
=
\mathbf{F}^{(1)}
\!\left(
\mathbf{x}_k^{(1)},
\mathbf{u}_k^{(1)}
\right),
&& k\in \mathbb{Z}_{1:N-1},
\label{eq:ocp_stg1_dynamics}\\
& \|
\mathbf{u}_k^{(1)}
\|_\infty
\le
f_{\max},
&& k\in \mathbb{Z}_{1:N-1},
\label{eq:ocp_stg1_control} \\
& s_d(\mathbf{r}_{c,k},\mathcal{P}^{t}_k) \ge \delta_{\mathrm{safe}, k},
&& k\in \mathbb{Z}_{1:N-1}.
\label{eq:ocp_stg1_koz}
\end{align}
\end{subequations}
}
\normalsize
\noindent The objective of the problem is to minimize the weighted sum of the control effort with a control-rate smoothness penalty. 
Eq.~\eqref{eq:ocp_stg1_dynamics} defines the discrete-time translational dynamics of the system COM, where Eq.~\eqref{eq:eom_discrete_nonlin} reduces to the HCW equation with a zero-order hold control input. 
The boundary conditions and the maximum control input are defined in Eqs.~\eqref{eq:ocp_stg1_bc} and ~\eqref{eq:ocp_stg1_control}, respectively. 
Finally, Eq.~\eqref{eq:ocp_stg1_koz} defines the safety constraint by enforcing a minimum clearance $\delta_{\mathrm{safe},k}$ between the servicer and the convex hull of the tumbling target via a signed-distance function.
Here, $\mathcal{P}^{t}_k$ denotes the time-varying keep-out geometry of the tumbling target, including its attitude and position, and $s_d(r_{c,k},\mathcal{P}^{t}_k)$ denotes the signed distance from the servicer's system COM to this set. 
Details of this implementation are available in \cite{virgili2019convex}. 
In particular, the minimum keep-out radius is defined in two phases: the first corresponds to the maximum extension length from the system COM, and the second to the radius implied by the servicer's final joint configuration.

\vspace{0.25em}
\subsubsection{Internal torque reconfiguration}

Given the optimal translational trajectory $\{\mathbf{x}_k^{\star(1)}, \mathbf{u}_k^{\star(1)}\}$ from Stage 1, Stage 2 optimizes the remaining internal torque allocation while requiring the system COM to evolve along the prescribed trajectory, without introducing additional translational control forces.
The decision variables are 
$
   \mathbf{x}_k^{(2)}
   =
   \big[
   \mathbf{q}_{k}^{\top},
   \boldsymbol{\alpha}_{k}^{\top},
   \boldsymbol{\omega}_{k}^{\top},
   \dot{\boldsymbol{\alpha}}_{k}^{\top}
   \big]^{\top}
   \in \mathbb{R}^{7 + 2 N_l}$ 
and 
$
   \mathbf{u}_k^{(2)}
   =
   \big[
   \boldsymbol{\tau}_{0,k}^{\top},
   \boldsymbol{\tau}_{m,k}^{\top}
   \big]^{\top}
   \in \mathbb{R}^{3 + N_l}.
$
Accordingly, Stage 2 solves
{\footnotesize
\begin{subequations}
\label{eq:ocp_rot}
\begin{align}
\min \quad 
& \sum_{k=1}^{N-1}
{\mathbf{u}_k^{(2)}}^{\!\top}
\mathbf{W}_3 \,
\mathbf{u}_k^{(2)} \label{eq:ocp_stg2_obj} 
\\
\mathrm{s.t.}\quad
& \mathbf{x}_1^{(2)} = \mathbf{x}_{\mathrm{init}}^{(2)}, \label{eq:ocp_stg2_dyn} \\
& \mathbf{x}_{k+1}^{(2)}
=
\mathbf{F}^{(2)}
\!\left(
\mathbf{x}_k^{(2)},
\mathbf{u}_k^{(2)};
\mathbf{x}_k^{\star(1)},
\mathbf{u}_k^{\star(1)}
\right), 
&& k\in\mathbb{Z}_{1:N-1},  \label{eq:ocp_stg2_dyn}\\
& \|\boldsymbol{\tau}_{0,k}\|_{\infty}
\le
\tau_{0,\max}, \
\|\boldsymbol{\tau}_{m,k}\|_{\infty}
\le
{\tau}_{m,\max}, 
&& k\in\mathbb{Z}_{1:N-1},  \label{eq:ocp_stg2_control}\\
& - \boldsymbol{\alpha}_{\max}
\le
\boldsymbol{\alpha}_{k}
\le
\boldsymbol{\alpha}_{\max},
&& k\in\mathbb{Z}_{1:N},  \label{eq:ocp_stg2_alpha} \\
& \hat{\mathbf{r}}^{\top}_{d,k} R(\mathbf{q}_k) \hat{\mathbf{b}}^{\mathrm{body}}
\ge \cos\theta_{\mathrm{fov}}, \quad &&  k\in\mathbb{Z}_{1:N}, \label{eq:ocp_stg2_point} \\
& \boldsymbol{\alpha}_{k}
=
\boldsymbol{\alpha}^{\mathrm{preset}}_k,
\quad
\dot{\boldsymbol{\alpha}}_{k}
=
\dot{\boldsymbol{\alpha}}^{\mathrm{preset}}_k,
&& k \ge k_{\mathrm{preset}},  \label{eq:ocp_stg2_preset} \\
& \boldsymbol{\omega}_{N}
=
\boldsymbol{\omega}_{\mathrm{term}}, \label{eq:ocp_stg2_terminal_omega} \\
& \hat{\mathbf{r}}^{\top}_{g,N} R(\mathbf{q}_N) \hat{\mathbf{r}}^{\mathrm{body}}_{\mathrm{ee}}
= 1. \label{eq:ocp_stg2_terminal}
\end{align}
\end{subequations}
}
The objective is a weighted quadratic control cost. 
Eq.~\eqref{eq:ocp_stg2_dyn} is a reduced form of Eq.~\eqref{eq:eom_discrete_nonlin} conditioned on the optimized COM trajectory and corresponding control profile. 
Eq.~\eqref{eq:ocp_stg2_control} bounds the control input magnitude, and Eq.~\eqref{eq:ocp_stg2_alpha} defines the admissible joint angle range. 
Eq.~\eqref{eq:ocp_stg2_point} enforces a pointing constraint, where the capture point must remain within the FOV cone centered on the surface normal of the face on which the arm is mounted (cf.\ green cone in Fig.~\ref{fig:conops}). 
Here, $\hat{\mathbf{r}}^{d,k}$ is the servicer-to-capture-point unit vector and $\hat{\mathbf{b}}^{\mathrm{body}}$ is the body-fixed boresight unit vector.
Eqs.~\eqref{eq:ocp_stg2_preset} and~\eqref{eq:ocp_stg2_terminal_omega} prescribe the terminal joint profile and bus angular velocity, with the joint angles and rates constrained over the final steps to suppress rapid motion near capture.
Finally, Eq.~\eqref{eq:ocp_stg2_terminal} imposes the terminal attitude condition requiring the EE approach direction to be perpendicular to the capture point surface; $\hat{\mathbf{r}}_{g,N}$ denotes the desired approach direction computed from the target's tumbling state, $\hat{\mathbf{r}}^{\mathrm{body}}_{\mathrm{ee}}$ denotes the EE approach direction in the body frame, and $R(\mathbf{q})$ maps from the servicer body frame to the target-centered RTN frame.
Note that the terminal COM position in Eq.~\eqref{eq:ocp_stg1_bc} is selected so that, for the prescribed final joint configuration, alignment of the EE approach direction in Eq.~\eqref{eq:ocp_stg2_terminal} places the EE at the capture point.

\subsection{SCP and Heuristic Initialization} \label{sec:method_initialization}

The two subproblems are cast as NLPs and solved via SCP \cite{mao2016successive, oguri2023successive}.
At each iteration, SCP minimizes an augmented-Lagrangian objective that penalizes violations of the nonconvex constraints, subject to convexified constraints and a trust region constraint.

Since SCP requires a suitable warm-start, heuristic initialization procedures are provided for both stages. 
For Stage 1, the keep-out-zone constraint in Eq.~\eqref{eq:ocp_stg1_koz} is the only nonconvex constraint in the problem. 
A natural initialization can therefore be obtained by solving the corresponding convex problem with this safety constraint temporarily omitted. 
For Stage 2, a heuristic reference trajectory can be constructed by orienting the spacecraft bus to track the target line of sight, linearly interpolating the arm joints toward their terminal configuration, and recovering the bus angular velocity and joint rates by finite-differencing the resulting pose sequence.

When solving the SCP with a warm-start, the first convex subproblem may face artificial infeasibility due to a strict trust region constraint \cite{mao2016successive}. 
To avoid this, if the first convex subproblem is infeasible, the trust region radius is expanded until a feasible convex subproblem is obtained. 
This procedure leads to near-100\% SCP success across both heuristic and learning-based warm-starts (cf. Sec.~\ref{sec:results}). 
Furthermore, the nonlinear dynamics are propagated using forward Euler integration in Stage~2.
Since this integration does not exactly preserve the unit-quaternion constraint, the quaternion is re-normalized before each nonlinear constraint evaluation and linearization in the SCP iteration.

\subsection{Conditional Behavior Cloning with Sequence Model}

In this paper, the proposed CT-based warm-starting is applied only to the second-stage attitude–manipulator optimization. 
This choice follows from the structure of the two-stage formulation; Stage 1 has linear HCW dynamics and a lower-dimensional state--control space, whereas Stage 2 involves nonlinear coupled bus–arm dynamics and repeated Jacobian computation inside SCP. 
Therefore, the computational bottleneck arises primarily in Stage 2.

To generate a high-quality warm-start, this paper uses sequence modeling for trajectory generation. 
Although sequence-modeling methods such as decision transformer (DT) have been applied to aerospace trajectory generation by conditioning on \textit{reward-to-go} \cite{chen2021decision} or \textit{constraint-to-go} \cite{art_ieeeaero24}, this formulation is not directly suitable for the space manipulator problem considered here. 
In problems where the dominant nonconvexity stems from nonlinear coupled dynamics, as in Eq.~\eqref{eq:ocp_rot}, constructing a meaningful reward-to-go signal from a relaxed lower-bound problem requires an additional NLP solve and is generally impractical. 
Moreover, because the system is highly constrained, optimization-based data generation is the only practical means of obtaining physically meaningful trajectories. 
As a result, the available dataset is dominated by expert demonstrations (i.e., locally optimal trajectories) rather than trajectories spanning a broad range of rewards and constraint violations. 
The trajectory generation problem is therefore formulated as conditional behavior cloning, where the model predicts expert state--control trajectories conditioned on the task specification.

The tokenized trajectory sequence is expressed as 
$\mathcal{T}_{1:N} = \{ \mathbf{o}_k, \mathbf{s}_k,  \mathbf{a}_k\}_{k=1}^{N}$, where $\mathbf{o}_k$, $\mathbf{s}_k$, and $\mathbf{a}_k$ denote the observation, state, and action, respectively. 
Accordingly, this paper defines the trajectory tokenization as 
\begin{align} \small
    \mathcal{T}_{1:N} = \left\{ttg, \{\mathbf{q}_{\mathrm{targ},k}, \boldsymbol{\omega}_{\mathrm{targ},k} \}, \{\mathbf{x}_k^{\star(1)}, \mathbf{u}_k^{\star(1)}\},\mathbf{x}_k^{(2)},  \mathbf{u}_k^{(2)} \right\}_{k=1}^{N}, 
\end{align}
where $ttg$ is the \textit{time-to-go}, which expresses the remaining time to the final epoch; 
$\{  \mathbf{q}_{\mathrm{targ},k}, \boldsymbol{\omega}_{\mathrm{targ},k} \}$ denotes the tumbling target's attitude state resolved in the target-centered RTN frame;
and $\{ \mathbf{x}_k^{\star(1)}, \mathbf{u}_k^{\star(1)}\}$ denotes the optimized system COM's translational trajectory in Stage 1. 

As an additional sequential-prediction variant, action chunking \cite{zhao2023learning} is considered in this work.
Instead of predicting only the immediate control input, action chunking predicts an $H$-step action sequence, which can improve temporal consistency over the rollout and has shown strong empirical performance in modern robotics \cite{zhao2023learning}. 
For the single-step decoder, the action token is defined as $\mathbf{a}_k = u^{(2)}_k$, whereas for action chunking with chunk length $H$, it is defined as $\mathbf{a}_k = [u^{(2)\top}_k, u^{(2)\top}_{k+1}, ..., u^{(2)\top}_{k+H-1} ]^{\top}$.

\begin{figure*}[th!]
     \centering
     \begin{subfigure}[th!]{0.4\linewidth}
         \centering
         \includegraphics[width=\linewidth]{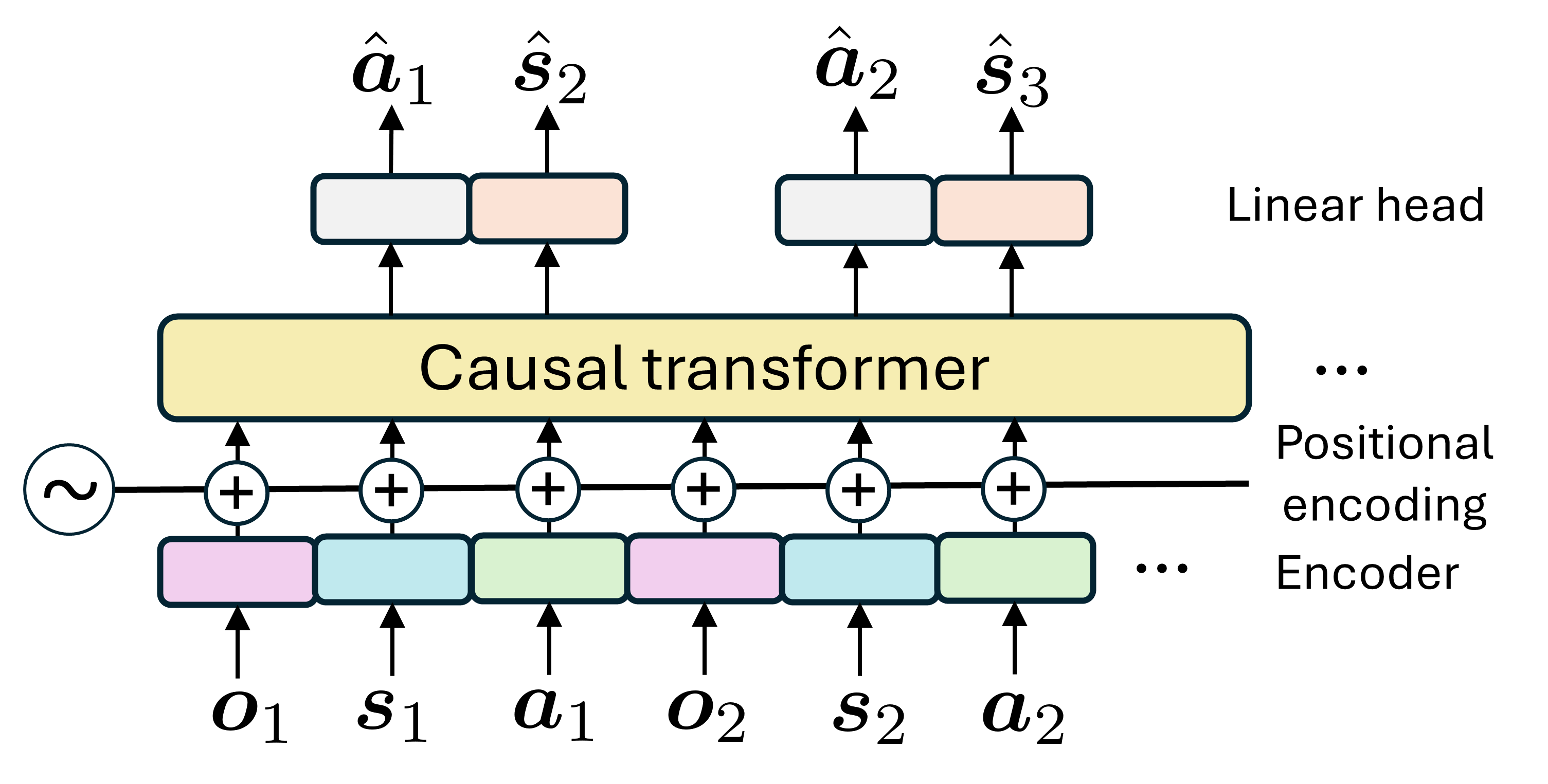}
         \caption{Linear Decoder}
         \label{fig:ct_linear_head}
    \end{subfigure} 
    \begin{subfigure}[th!]{0.4\linewidth}
         \centering
         \includegraphics[width=\linewidth]{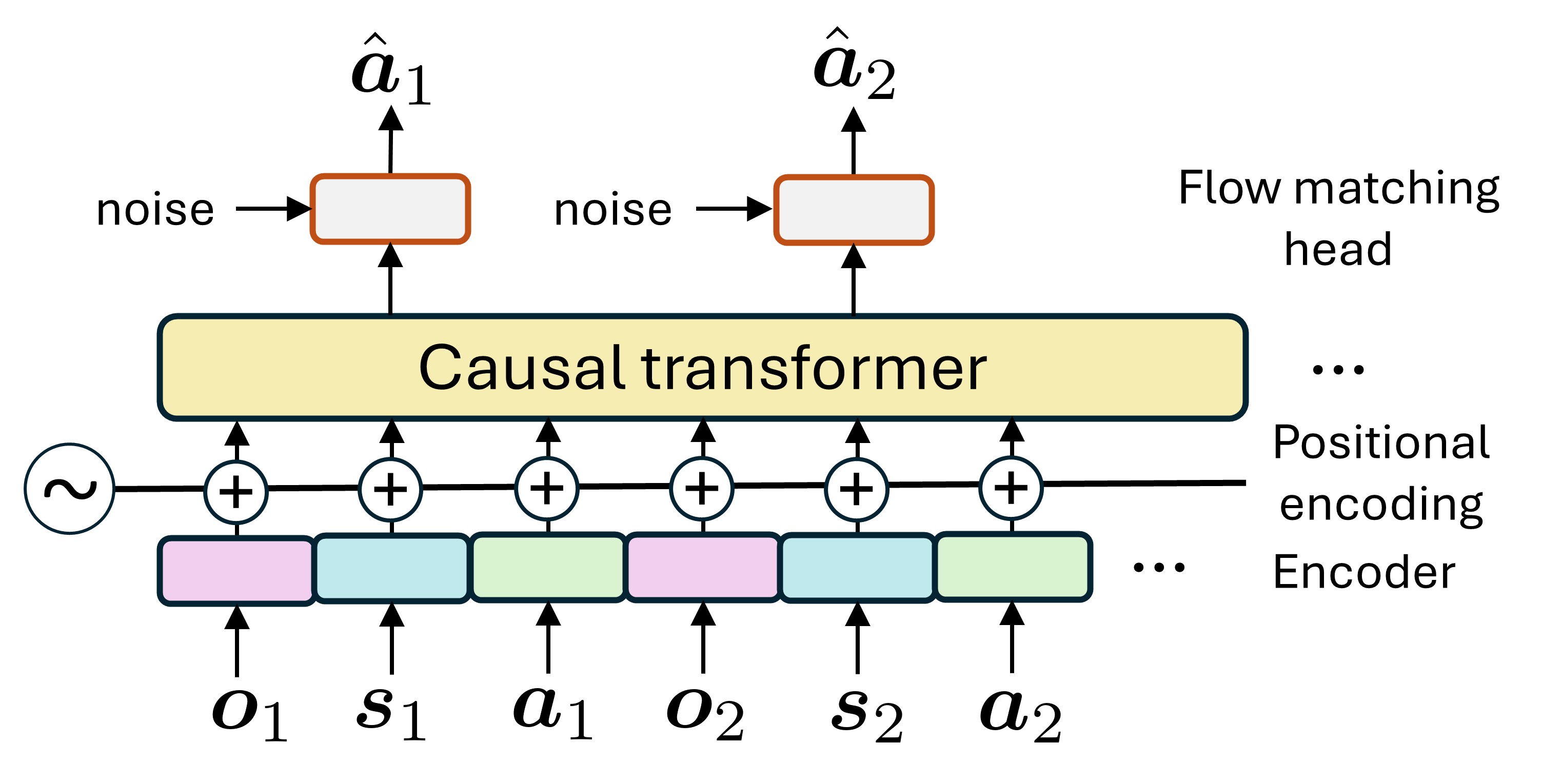}
         \caption{Flow Matching Decoder}
         \label{fig:ct_flow_matching}
     \end{subfigure}
    \caption{Variants of the causal transformer (CT) architectures.}
    \label{fig:causal_transformer}
\end{figure*}

Two decoder architectures are considered in this paper, as summarized in Fig.~\ref{fig:causal_transformer}.

\subsubsection{Linear Decoder}
In the original implementation of DT \cite{chen2021decision}, a linear decoder maps the CT latent representation to the physical control input:
\begin{align}
    \hat{\mathbf{a}}_k = W_\theta h_k + b_\theta,
\end{align}
where $W_\theta$ and $b_\theta$ are the learnable parameters, and $h_k$ is the latent output from the CT.  
During training, the loss function is set to a batched sum-of-squares loss: 
\begin{align}
    \mathcal{L}_{\text{Lin}} =  
 \sum_{b=1}^{B} \sum_{k=1}^{N} {\|\hat{\mathbf{x}}_k^{(b)} - \mathbf{x}_{\mathrm{GT},k}^{(b)}\|^2 
 +
 \|\hat{\mathbf{a}}_{k}^{(b)} - \mathbf{a}_{\mathrm{GT},k}^{(b)}\|^2 }, 
\end{align}
where $\{\hat{\mathbf{x}}_k^{(b)}, \hat{\mathbf{a}}_k^{(b)} \}$ are the predicted state and action, and the $\{\mathbf{x}_{\mathrm{GT},k}^{(b)}, {\mathbf{a}}_{\mathrm{GT},k}^{(b)} \}$ are the ground-truth state and action sampled from the training data. 

\subsubsection{Flow Matching Decoder}
Recently, vision--language--action (VLA) models have increasingly adopted diffusion- and flow-based action decoders, motivated by their ability to represent nonlinear and multimodal continuous action distributions \cite{black2024pi_0, lipman2022flow}. 
In contrast to an affine action head that directly maps the transformer's latent output to an action, FM learns a conditional velocity field that transports samples from a simple prior distribution to the action distribution in the dataset.

During training, FM defines an interpolant 
\begin{align}
    z_\tau = (1-\tau)z_0 + \tau \mathbf{a}_{\mathrm{GT},k}, \quad z_0 \sim \mathcal{N}(0,I), \quad \tau \in [0,1],
\end{align}
and trains a conditional vector field $v_\theta(z_\tau,\tau,h_k)$ to match the corresponding transport velocity using the batched loss function 
\begin{align} \footnotesize
    \mathcal{L}_{\mathrm{FM}}
    =
    \sum_{b=1}^{B}  \sum_{k=1}^{N}
    \left\|
    v_\theta(z_{\tau}^{(b,k)},\tau,h_{k}^{(b)})
    -
    \left(\mathbf{a}_{\mathrm{GT}, k}^{(b)} - z_{0}^{(b,k)}\right)
    \right\|^2 .
\end{align}
At inference, an action sample is generated by initializing $z(0)\sim\mathcal{N}(0,I)$ and integrating the learned ordinary differential equation (ODE)
\begin{align}
    \frac{d z(\tau)}{d\tau}
    =
    v_\theta(z(\tau),\tau,h_k),
    \qquad
    \hat{\mathbf{a}}_k = z(1).
\end{align}
The following section evaluates how the number of ODE integration steps affects inference time and warm-start quality.

For both decoders, the trajectory is propagated with the available dynamics (cf. Eq.~\eqref{eq:ocp_stg2_dyn}) using the generated action history, thereby making the warm-start dynamically feasible. 

\section{Results} \label{sec:results}

This section evaluates the proposed learning-based warm-start on a terminal approach case study for a space manipulator. \footnote{All scripts are implemented in Python using JAX-based automatic differentiation and just-in-time (JIT) compilation. Computations are performed on an NVIDIA Jetson AGX Orin 64 GB (Arm Cortex-A78AE v8.2 64-bit CPU and an NVIDIA Orin GPU).}
The experiments address three questions: (i) whether the learned warm-start reduces Stage 2 SCP iterations and runtime without degrading final solution quality; (ii) how decoder architecture, action chunking, and training dataset size affect warm-start performance; and (iii) how the learned warm-start affects the trade-off between computational effort and realized control cost when SCP is terminated at feasibility rather than local optimality.

\subsection{Implementation}

The configuration of the spacecraft servicer and tumbling object is summarized in Table~\ref{tab:spacecraft_config}. 
The servicer is equipped with a three-link arm ($N_l=3$) with revolute joints.
The tumbling spacecraft is in a low Earth orbit and has dimensions of $80 \times 80 \times 40$ cm, with the capture point at the center of the $+z$ face. 
The total transfer time is set to $t_f = 1{,}200$~s, with $N=60$ time steps.

\begin{table}[t]
\centering
\footnotesize
\caption{Servicer and target spacecraft configuration.}
\label{tab:spacecraft_config}
\setlength{\tabcolsep}{2.5pt}
\renewcommand{\arraystretch}{0.92}
\begin{tabular}{@{}lp{0.33\columnwidth}clp{0.3\columnwidth}@{}}
\hline
\multicolumn{2}{c}{\textbf{Servicer}} & & 
\multicolumn{2}{c}{\textbf{Target}} \\
\cline{1-2} \cline{4-5}
Property & Value & & Property & Value \\
\hline
Mass
& $500$ kg
& & Mass
& $50$ kg \\

Dim.
& $1.0{\times}1.0{\times}1.0$ m
& & Dim.
& $0.8{\times}0.8{\times}0.4$ m \\

Inertia
& $(83.3,83.3,83.3)$ kg\,m$^2$
& & Inertia
& $(2.5,3.0,2.0)$ kg\,m$^2$ \\

Link mass
& $10$ kg each
& &  $\hat{\mathbf{b}}^{\mathrm{body}}$
& $[0,0,1]^\top$  \\

Link length
& $\{0.75,0.75,0.30\}$ m
& & Tumble 
& $\leq10^{-3}$ rad/s per axis \\
$\boldsymbol{\alpha}_{\max}$
& $[\pi, \pi/2, 1.75]^\top$ rad
& & Attitude
& random SO(3) \\
$f_{\max}$
& $2.5$ N
& & Orbit ($a$) 
& $6,800$ km \\
$\tau_{0,\max}, \tau_{m,\max}$
& $5\cdot 10^{-3}$ N\,m
& & Orbit ($e$) &  $10^{-3}$
\\
\multirow{2}{*}{$\delta_{\mathrm{safe},k}$} & $
 1.85 \ \text{m} \ (t_k \leq 0.7 \ t_f)$ &  & Orbit ($i$) & $97^\circ$
\\
\cline{4-5}
&$ 1.46 \ \text{m} \ (0.7 \ t_f < t_k)$&& $\mathbf{W}_1, \mathbf{W}_2$& $\mathbf{I}_3, \ 0.03 \ \mathbf{I}_3$  \\
$\theta_{\mathrm{fov}}$& $20^\circ$ && $\mathbf{W}_3$ & $\mathrm{diag}(0.5 \ \mathbf{I}_3, \mathbf{I}_3)$ \\
\hline
\end{tabular}
\end{table}

\begin{figure*}
    \centering
    \includegraphics[width=1.0\linewidth]{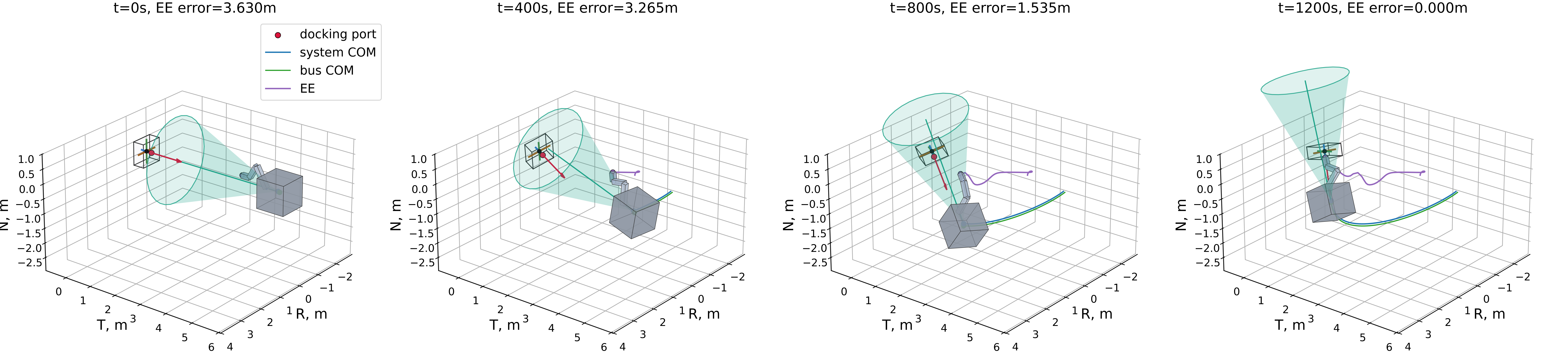}
    \caption{Representative terminal approach trajectory to the tumbling target, shown in the target-centered RTN frame.  }
    \label{fig:rep_traj_3d}
\end{figure*}

To generate a training dataset, the two-stage SCP with the heuristic warm-start strategy is employed, and only the locally optimal solutions are collected. 
To promote generalization, each training scenario is generated by independently sampling the initial conditions from uniform distributions.
The servicer is initialized at a nominal $5\,\mathrm{m}$ standoff along the line-of-sight to the capture point to avoid unrealistic and ill-posed geometry, with a positional offset drawn from $\delta r_{\mathrm{RTN}} \sim \mathrm{Unif} \ [\pm1, \pm1, \pm 2]$ m, and an initial RTN velocity sampled per axis as $v_{0,i} \sim \mathrm{Unif}[-5\cdot 10^{-3}, 5\cdot 10^{-3}]\,\mathrm{m/s}$.
The servicer's initial attitude is set deterministically to point toward the target center of mass, and the arm starts from a fixed joint configuration 
$\mathbf{\alpha}_{0} = (-\pi/9,\ \pi/5,\ 0)\,\mathrm{rad}$.
Fig.~\ref{fig:rep_traj_3d} illustrates a sample trajectory that is optimized with the two-stage SCP pipeline.  

As discussed in Sec.~\ref {sec:methodology}, this paper applies the learning-based warm-start only to Stage 2 SCP. 
To empirically verify this bottleneck, the stage-wise runtime across 300 held-out test scenarios using the heuristic two-stage SCP pipeline is first measured.  
The Stage 1 solve required $0.44 \pm 0.15$~s (mean $\pm$ std.), whereas the Stage 2 solve required $4.20 \pm 2.03 $~s.\footnote{Each convex subproblem was canonicalized as a disciplined parameterized program (DPP), where one-time canonicalization cost is excluded from the reported runtime.} 
Therefore, Stage 2 accounts for approximately $90\%$ of the total solve time, making it the dominant computational bottleneck and motivating the focus on Stage 2 warm-starting.

\subsection{Flow Steps in Flow Matching Decoder}

Before comparing the two decoder architectures, a sensitivity analysis is conducted on the number of forward-Euler integration steps used by the FM decoder over the normalized interval $\tau \in [0,1]$, denoted by $n_{\mathrm{flow}}$. Table~\ref{tab:nflow_sweep} reports this sensitivity study using the same FM model trained on 90k trajectories and single-step action prediction ($H=1$). The metrics include inference time, terminal EE position error relative to the capture point, subsequent SCP iterations, and SCP success rate. 
Values are reported as the mean $\pm$ std across the 300 held-out test scenarios. 

The results show that inference time scales approximately linearly with $n_{\mathrm{flow}}$ due to the increased number of velocity-field evaluations. 
However, larger $n_{\mathrm{flow}}$ does not improve warm-start quality, and  $n_{\mathrm{flow}}=1$ and $2$ achieve the lowest terminal errors and are among those requiring the fewest SCP iterations while maintaining nearly identical success rates. 
This suggests that excessive flow integration can accumulate errors from the learned velocity field through nonlinear propagation, increasing inference cost without meaningful performance gain. 
This behavior is also consistent with the dataset being generated by a deterministic optimization pipeline, which makes the history-conditioned action distribution nearly unimodal.
In the following experiments, $n_{\mathrm{flow}}=1$ is used as a baseline for the FM decoder. 

\begin{table}[t!]
\centering
\footnotesize
\caption{Flow matching ODE steps ($n_{\text{flow}}$) sweep for the same trained model.}
\label{tab:nflow_sweep}
\begin{tabular}{rcccc}
\hline
$n_{\mathrm{flow}}$ & \makecell[c]{Inference \\ time [s]} & \makecell[c]{EE position \\ error [m]} & \makecell[c]{SCP \\iterations} & \makecell[c]{SCP \\ success rate [\%]} \\
\hline
1 & $0.76 \pm 0.15$ & $0.14 \pm 0.08$ & $13.8 \pm 6.5$ & 99.7 \\
2 & $1.43 \pm 0.16$ & $0.14 \pm 0.12$ & $13.5 \pm 7.1$ & 99.7 \\
4 & $2.74 \pm 0.18$ & $0.21 \pm 0.23$ & $14.6 \pm 8.2$ & 100.0 \\
8 & $5.40 \pm 0.22$ & $0.25 \pm 0.22$ & $15.5 \pm 8.7$ & 99.7 \\
10 & $6.50 \pm 0.28$ & $0.25 \pm 0.22$ & $15.0 \pm 8.6$ & 99.3 \\
\hline
\end{tabular}
\end{table}

\subsection{Comparison of Decoder Options}

\begin{table*}[t!]
\centering
\caption{Comparison of training configurations across decoder types and chunk lengths.}
\begin{tabular}{lcc|ccccc}
\hline
Method & Decoder & Chunk Length & SCP success rate [\%] & Cost &
SCP iterations & Stage 2 runtime [s] & Inference time [s] \\
\hline
Heuristic & -- & -- & 99.3 & $0.327 \pm 0.115$ & $20.45 \pm 9.66$ & $4.20 \pm 2.03$ & -- \\
\hline
CT & Linear & 1 & 99.7 & $0.324 \pm 0.112$ & $\mathbf{15.20 \pm 6.33}$ & $\mathbf{3.80 \pm 1.35}$ & $0.68 \pm 0.01$ \\
CT & Linear & 5 & 99.7 & $0.324 \pm 0.112$ & $\mathbf{19.86 \pm 7.79}$ & $4.78 \pm 1.65$ & $0.71 \pm 0.01$ \\
CT & FM     & 1 & 99.7 & $0.323 \pm 0.112$ & $\mathbf{13.82 \pm 6.52}$ & $\mathbf{3.54 \pm 1.31}$ & $0.76 \pm 0.15$ \\
CT & FM     & 5 & 100.0 & $0.324 \pm 0.112$ & $\mathbf{17.07 \pm 7.48}$ & $\mathbf{3.71 \pm 1.54}$ & $0.23 \pm 0.15$ \\
\hline
\end{tabular}
\label{tab:summary}
\end{table*}

As the primary evaluation of the CT-based behavior-cloning warm-start, this subsection addresses two aspects: performance with the full training dataset and performance scaling as the amount of training data is reduced.

\subsubsection{Decoder Architecture and Action-Chunk Length} 
Table~\ref{tab:summary} summarizes the comparative results obtained using the same training dataset, consisting of 90k feasible and locally optimal trajectories, across different decoder architectures and action chunk lengths ($H=1$ and $H=5$). 
The evaluation is conducted on 300 unseen in-distribution scenarios, with the optimization objective set identical to that of the original Stage 2 problem. 
Accordingly, these held-out scenarios evaluate generalization across unseen initial conditions within the prescribed sampling distribution, rather than robustness to state-estimation uncertainty or uncertainty in the target’s tumbling state and model parameters.
For action-chunked control generation, the linear decoder adopts a receding-horizon predict-$H$/execute-1 strategy, in which only the first action in each predicted chunk is applied before replanning at the next time step. 
In contrast, the FM decoder applies the predicted $H$-step action sequence open-loop before replanning, corresponding to a predict-$H$/execute-$H$ strategy.
At inference, the linear decoder uses KV caching to reduce runtime.\footnote{KV caching reduced the linear decoder's runtime by a few percent. For FM, however, the latent input to the decoder changes at every integration step. 
This requires repeated transient queries and cache updates, whose overhead exceeded the saved computation time.}
Also note that the Stage 2 runtime listed in the table includes the inference time. 

First, the SCP success rate and final cost distribution are nearly identical across all five cases, including the heuristic warm-start described in Sec.~\ref{sec:method_initialization}. 
With sufficient training data, both decoder variants reduce the SCP iteration count relative to the heuristically warm-started SCP, by $26\%$ (linear) and $32\%$ (FM).
These reductions translate into $10\%$ and $16\%$ decreases in the total Stage 2 runtime with reduced variance, respectively, even after accounting for the network's inference time.
Under the tested configuration, the FM decoder yields fewer SCP iterations and lower total Stage 2 runtime than the corresponding linear decoder.
Interestingly, increasing the action chunk length $H$ does not improve the quality of the learned trajectory as a warm-start, despite the reduction in inference time with the FM decoder. 
In direct robot-policy deployment, action chunking can improve temporal consistency by smoothing the executed control sequence \cite{zhao2023learning}. 
However, generated trajectories are only used to initialize SCP rather than being executed directly. 
The SCP serves as a trajectory-level correction step, enforcing dynamics and constraints while regularizing the control history through the quadratic control objective. 
Thus, longer open-loop chunks provide limited benefit, whereas single-step prediction can exploit the most recent trajectory history at every step.
Overall, the CT-based warm-start reaches the same final cost distribution as the two-stage heuristic baseline while reducing the Stage 2 SCP iteration count, thereby reducing the overall runtime.

\subsubsection{Training-data Scaling} 
Having established the full-data performance, the following question is how the two decoder architectures degrade as the available training set is reduced.
Here, the two decoder architectures (linear and FM) are trained with $N_{\mathrm{data}}=\{5\text{k},10\text{k},90\text{k}\}$ trajectories. 

\begin{figure}[th!]
     \centering
     \begin{subfigure}[th!]{0.49\linewidth}
         \centering
         \includegraphics[width=\linewidth]{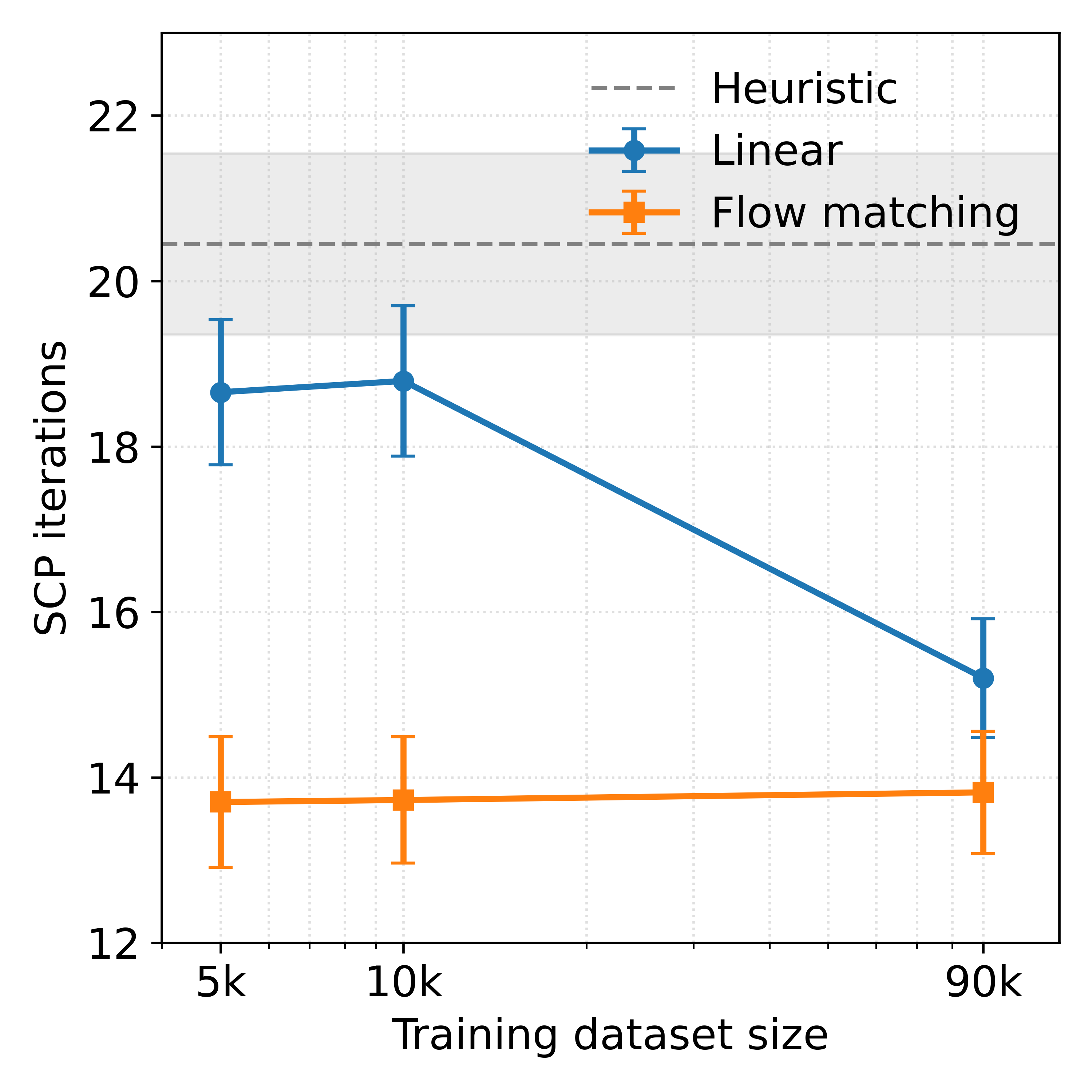}
         \caption{SCP iterations}
         \label{fig:sample_efficiency_iters}
    \end{subfigure} 
    \begin{subfigure}[th!]{0.49\linewidth}
         \centering
         \includegraphics[width=\linewidth]{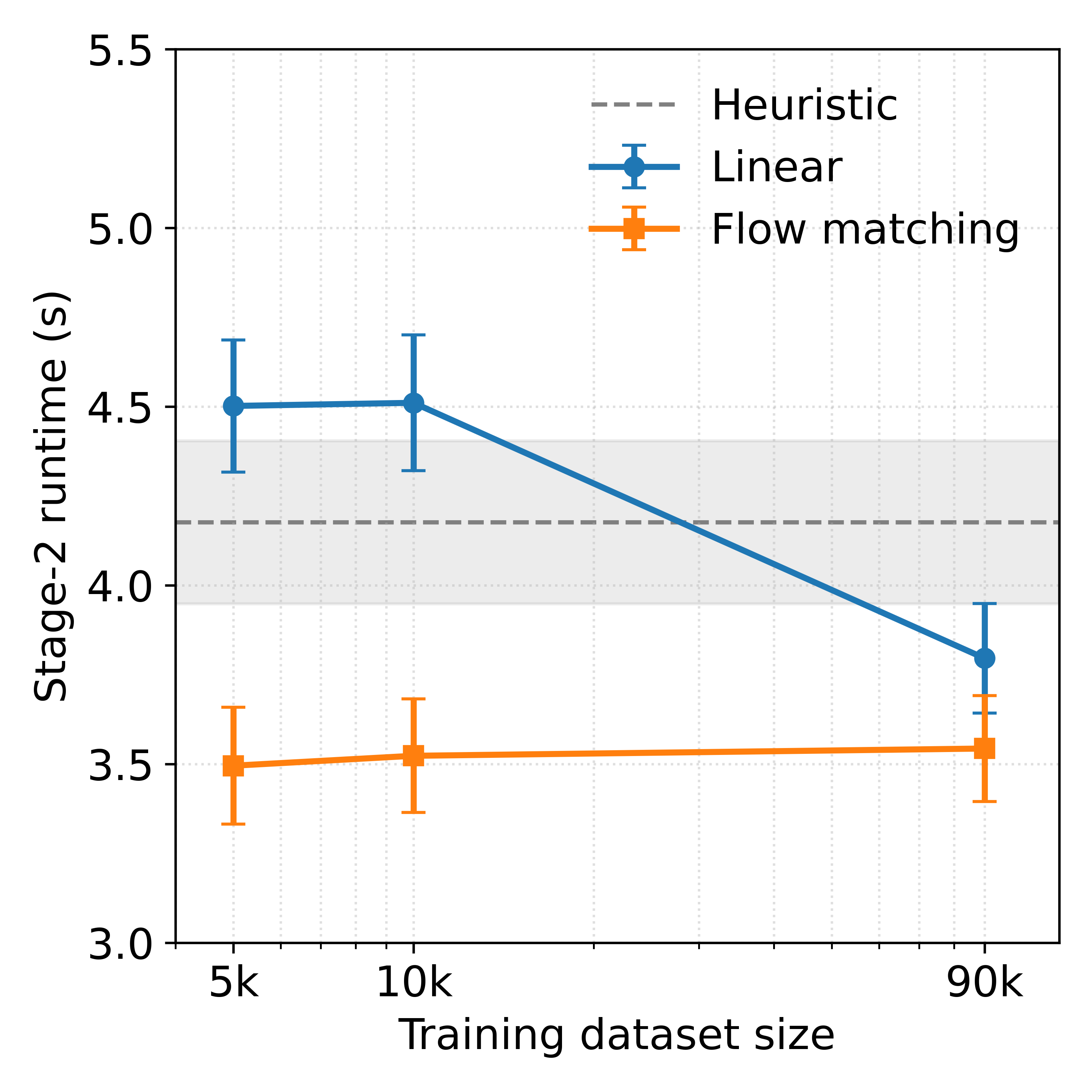}
         \caption{Total runtime}
         \label{fig:sample_efficiency_runtime}
     \end{subfigure}
    \caption{Training-data scaling across decoder architectures. }
    \label{fig:sample_efficiency}
\end{figure}

\begin{table*}[t!]
\centering
\caption{Comparison of feasibility- and optimality-based SCP under heuristic and CT warm-starts.}
\label{tab:feas-vs-cost}
\begin{tabular}{ll c c c | c c c}
\hline
Problem & Warm-start & SCP success rate [\%] &
Mean SCP iterations & Mean runtime [s] &
Cost (mean$\pm$std) & Cost (median) & Cost (p99) \\
\hline
\multirow{2}{*}{Optimality}
 & Heuristic & 99.3 &
 $20.45$ & $4.20$ &
 $0.327 \pm 0.115$ & $0.290$ & $0.681$ \\
 & CT/FM/1 & 99.7 &
 $\mathbf{13.82}$ & $\mathbf{3.54}$ &
 $0.323 \pm 0.112$ & $0.290$ & $0.658$ \\
\hline
\multirow{2}{*}{Feasibility}
 & Heuristic & 99.7 &
 $\mathbf{10.73}$ & $\mathbf{1.22}$ &
 $1.338 \pm 2.323$ & $\mathbf{0.685}$ & $12.161$ \\
 & CT/FM/1 & 100.0 &
 $12.50$ & $2.16$ &
 $\mathbf{0.761 \pm 0.252}$ & $0.727$ & $\mathbf{1.616}$ \\
\hline
\end{tabular}
\end{table*}

Fig.~\ref{fig:sample_efficiency} summarizes the SCP iteration count and total Stage 2 runtime, including model inference, for the six trained configurations described above. 
Across all configurations, SCP success exceeds \(99\%\), with comparable final-cost distributions, consistent with Table~\ref{tab:summary}. 
Error bars denote the 95\% confidence interval for the mean, while the dashed line and gray band indicate the mean of the heuristic baseline and its 95\% confidence interval from Table~\ref{tab:summary}.

The results show that the FM decoder maintains approximately 14 SCP iterations across the 5k, 10k, and 90k training sets, indicating stronger training-data efficiency than the linear decoder. 
In contrast, the linear decoder degrades substantially as the training dataset is reduced, with its total runtime exceeding that of the heuristic SCP baseline at the smaller dataset sizes.
This behavior is consistent with the FM head providing a more flexible nonlinear action map from the transformer representation to the control sequence, which appears beneficial in the data-scarce regime.
However, because inference uses only $n_{\mathrm{flow}}=1$, the decoder is unlikely to fully exploit the multimodal sampling capability typically associated with flow-based models. 
This is consistent with the structure of the dataset, which is generated by a deterministic SCP pipeline and is therefore expected to induce a nearly unimodal conditional action distribution around each trajectory history. 
Thus, the practical benefit of FM in this case study is best interpreted as improved training-data efficiency through a lightweight nonlinear decoder.

\subsection{Feasibility Problem}

Although the experiments in the previous subsection minimize the quadratic control cost in Eq.~\eqref{eq:ocp_stg2_obj}, strict cost optimality is not always the primary requirement for onboard trajectory generation. 
In a flight implementation, particularly during terminal approach, obtaining a dynamically feasible and constraint-satisfying trajectory within a limited computation time can be more important than further reducing the control effort.
Motivated by this observation, this subsection considers a feasibility projection, in which the objective is set to zero \cite{takubo2026language, lampariello2013generating}. 
Since this formulation seeks only a constraint-satisfying trajectory rather than local optimality in control cost, SCP is expected to terminate faster than the cost-optimal SCP.

To this end, this subsection evaluates whether a learned warm-start retains its value when seeding this feasibility projector, using the same $300$ held-out scenarios.
The model trained on the 90k-trajectory dataset with the FM decoder and single-step action prediction is adopted as the representative CT model. 
Table~\ref{tab:feas-vs-cost} summarizes the results of the experiment, where the optimality rows coincide with the Heuristic and CT/FM/1 rows of Table~\ref{tab:summary}. 
Note that the statistics are computed over numerically feasible cases, and the reported cost is the realized coupled control effort, evaluated with the same function.

Solving the feasibility problem roughly halves the SCP iterations for the heuristic warm-start and reduces the Stage 2 runtime for both warm-start methods, but the resulting trajectories differ sharply.
Since the solver terminates at the first feasible iterate rather than at a local cost optimum, the realized cost is largely determined by the basin reached by the initialization. 
For the heuristic seed, this shortcut is unreliable: although the median cost remains modest, the upper tail grows dramatically, with the 99th-percentile cost increasing from $0.68$ under optimality-based termination to $12.16$ under feasibility-based termination. 
In contrast, under feasibility-based termination, the CT warm-start substantially suppresses the high-cost tail relative to heuristic initialization, reducing the 99th-percentile cost to 1.62 with a 100\% SCP success rate. 
This improvement comes with an increase in mean runtime from $1.22$ to $2.16$~s relative to heuristic initialization, while remaining below the $3.54$~s required by the CT warm-start under optimality-based termination.

\section{Conclusion} \label{sec:conclusion}

This paper presents a learning-based warm-starting framework for sequential convex programming (SCP)-based terminal approach guidance of a space manipulator to a tumbling object. 
Applying a causal-transformer (CT)-based warm-start to the coupled attitude–manipulator stage of a two-stage trajectory-optimization pipeline reduces SCP iterations and runtime while preserving the final-cost distribution of the heuristic baseline. 
The decoder ablation shows that both linear and flow matching (FM) heads provide effective warm-starts; FM is more robust to reductions in training-set size, whereas action chunking does not improve subsequent SCP convergence. 
When SCP terminates upon feasibility, the learned warm-start also suppresses the high-cost tail observed with heuristic initialization. 
Overall, these results demonstrate that learned sequence models can serve not only as fast trajectory predictors but also as trajectory priors that guide SCP toward basins containing feasible, low-cost solutions. 
Future work will extend the framework to account for uncertainty in the target’s tumbling state and model parameters.

\section*{APPENDIX}
Table \ref{tab_appendix:hyper} presents an overview of the CT's hyperparameter settings used in this work.
All variables are scaled to the range $[-1,1]$ using the component-wise minimum and maximum values of the warm-start trajectory to improve numerical conditioning in the optimization solver.

\begin{table}[ht]
\centering
\caption{Hyperparameters of CT.}
\begingroup
\renewcommand*{\arraystretch}{1.1}
\begin{tabular}{l l}
    \hline
     Hyperparameter & Value \\
    \hline
     Number of layers & 6\\
     Number of attention heads & 6 \\
    Embedding dimension & 384 \\
     Batch size& 4 \\
    Activation function & ReLU\\
    Dropout & 0.1\\
    Learning rate & $3\times10^{-5}$ (linear decay)\\
    Gradient norm clip & 1.0 \\
    Gradient accumulation iters & 8 \\
    \hline
    \end{tabular}%
    \label{tab_appendix:hyper}
    \endgroup
\end{table}

\vspace{-1em}
\section*{ACKNOWLEDGMENT}

The authors acknowledge financial support from the Ezoe Memorial Recruit Foundation (Yuji Takubo) and the Department of Defense National Defense Science and Engineering Graduate Fellowship (NDSEG; Maximilian Adang).

\bibliographystyle{ieeetr}
\bibliography{reference}

\end{document}